\documentclass[letterpaper]{article}
\usepackage{aaai25}  
\usepackage{times}  
\usepackage{amsmath}
\usepackage{amssymb}
\usepackage{helvet}  
\usepackage{courier}  
\usepackage[hyphens]{url}  
\usepackage{graphicx} 
\usepackage{natbib}  
\usepackage{caption} 
\DeclareCaptionStyle{ruled}{labelfont=normalfont,labelsep=colon,strut=off} 
\usepackage{algorithm}
\usepackage{algorithmic}
\usepackage{multirow}

\begin{document}
%
\title{HiHa: Introducing Hierarchical Harmonic Decomposition to Implicit Neural Compression for Atmospheric Data}
\author{Zhewen Xu, Baoxiang Pan, Hongliang Li, Xiaohui Wei}
\author{
	Zhewen Xu\textsuperscript{\rm 1},
	Baoxiang Pan\textsuperscript{\rm 2},
	Hongliang Li\textsuperscript{\rm 1}
	Xiaohui Wei\textsuperscript{\rm 1}
}
\affiliations {
	\textsuperscript{\rm 1}Jilin University, Changchun, China \\
	\textsuperscript{\rm 2}Institute of Atmospheric Physics, Beijing, China \\
	weixh@jlu.edu.cn, lihongliang@jlu.edu.cn
}
%
%

%
%

\maketitle
\begin{abstract}
\begin{quote}
The rapid development of large climate models has created the requirement of storing and transferring massive atmospheric data worldwide. Therefore, data compression is essential for meteorological research, but an efficient compression scheme capable of keeping high accuracy with high compressibility is still lacking. As an emerging technique, Implicit Neural Representation (INR) has recently acquired impressive momentum and demonstrates high promise for compressing diverse natural data. However, the INR-based compression encounters a bottleneck due to the sophisticated spatio-temporal properties and variability. To address this issue, we propose Hierarchical Harmonic decomposition implicit neural compression (HiHa) for atmospheric data. HiHa firstly segments the data into multi-frequency signals through decomposition of multiple complex harmonic, and then tackles each harmonic respectively with a frequency-based hierarchical compression module consisting of sparse storage, multi-scale INR and iterative decomposition sub-modules. We additionally design a temporal residual compression module to accelerate compression by utilizing temporal continuity. Experiments depict that HiHa outperforms both mainstream compressors and other INR-based methods in both compression fidelity and capabilities, and also demonstrate that using compressed data in existing data-driven models can achieve the same accuracy as raw data. The source code can be found at \url{https://anonymous.4open.science/r/HiHa-EB13/}.

\end{quote}
\end{abstract}

%
%

\section{Introduction}
Meteorology, a significant natural science, necessitates the analysis and integration of multiple atmospheric variables over hundreds of years \citep{bi2023accurate}.
Produced by meteorological organizations (e.g. ECMWF, CMA, JMA, etc.), petabytes of atmospheric data requires to be shared with researchers globally per day.
Therefore, the rapid development of meteorological research has spawned a huge demand for atmospheric data storage and transmission, while bringing about substantial costs citep{rasp2024weatherbench}.
This presents an enduring challenge, bringing huge stress to data storage and transmission \citep{han2024cra5}.
To alleviate the issue, many compression methods have been developed, such as BUFR, GRIB1, GRIB2 \citep{murrieta2015grib2}, NetCDF \citep{rew1990netcdf}, etc.
These lossless compressions are able to provide up to $4\times$ compression ratio, but still has $>$100 TB for 40 years' hourly data of 5 variables (basic data for training Pangu \citep{bi2023accurate}), which is insufficient in addressing the performance bottleneck of atmospheric data transmission \citep{han2024cra5}.

\begin{figure}[t]
	\centering
	\includegraphics[width=0.5\textwidth]{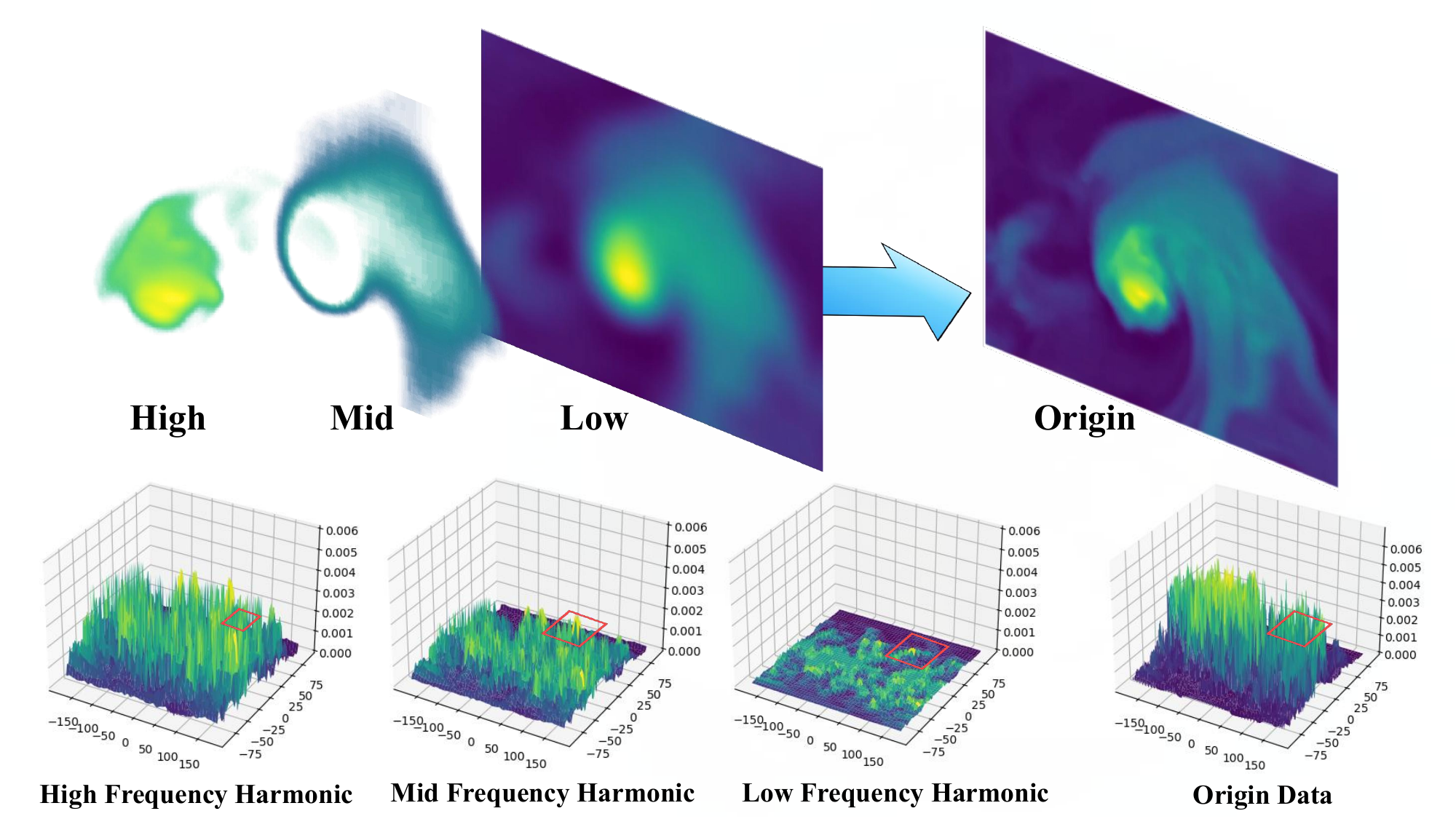}
	\caption{Harmonic decomposition of atmospheric data (an example of humidity in 500hpa).}
	\label{harmonic}
 \vspace{-8pt}
\end{figure}

\textbf{I}mplicit \textbf{N}eural \textbf{R}epresentation (\textbf{INR}), using periodic activation function \citep{sitzmann2020implicit}, has brought innovation to data representation, and possesses the ability to fit models accurately and rapidly in a concise manner \citep{2020NeRF}.
The INR-based compression technique, \textbf{I}mplicit \textbf{N}eural \textbf{C}ompression (\textbf{INC}), is gaining success on natural data compression ratio and enhanced spatial continuity, such as SCI \citep{yang2023sci}, MINER \citep{saragadamminer} and Gaussian Splatting \citep{zhang2024gaussianimage}, etc.
The inherent continuity in INR is highly compatible with the spectral characteristics of dynamic equations of atmospheric motion \citep{sitzmann2020implicit}.
In addition, given sufficient parameters, the universal approximation theorem of neural networks promises high fidelity representation \citep{gallant1988there}.
The above perspectives illustrate the potential of INR to generate an efficient and compact representation of atmospheric data.

Despite the high efficiency on other natural data \citep{yang2023tinc, kim2024c3}, the INC methods often encounter over-smoothness and low efficiency on dealing with atmospheric data.
As shown in Fig. \ref{harmonic}, the sophisticated spatio-temporal properties and variability caused by superposition of multiple complex harmonics leads to a high requirement of the fitting ability of neural networks\citep{sasamori1972linear}, and thereby causing the issue.
The issue can be further listed into the following crucial insights:
\begin{itemize}
\item Atmospheric data contain various harmonics and the fitting rules are sophisticated. The description of such processes within a confined parameter space poses significant challenges \citep{shin2024deep};
\item Spatial singularities will affect the searching process of the best parametric space as noisy bias, increasing the difficulty of the INR convergence \citep{acorn};
\item Due to the spherical topology of earth, employing Cartesian basis vector would result in the geographical information loss \citep{russwurm2023geographic};
\item The existing methods disregard the temporal continuity in atmospheric data, causing inefficiency in temporal data compression\citep{yang2023tinc};
\end{itemize}

Inspired by the above insights, we made statistical analysis of different harmonic components and propose \textbf{Hi}erarchical \textbf{Ha}rmonic decomposition implicit neural compression for atmospheric data (\textbf{HiHa}).
We explore the characteristics and spatial distribution of multiple harmonics in atmospheric data, and decompose the harmonic components into low, mid and high-frequency based on the harmonic frequency.
Leveraging spherical coordinates basis, we conduct a frequency-based hierarchical compression strategy including a multi-scale INR module, an iterative decomposition module and a sparse storage module on harmonics of low, medium and high frequency, hierarchically.
Moreover, we institute temporal residual compression module and utilizes a temporal multiscale Laplacian pyramid architecture for accelerating compression of atmospheric data over successive periods.

Comprehensive experimental results demonstrates that HiHa achieves impressive compression accuracy and efficiency compared with other baselines.
Specifically, at an atmospheric acceptable accuracy, HiHa achieves a compression ratio of over 200$\times$ within 43 seconds in error of 1e-3, surpassing other INC baselines in terms of both speed and accuracy.
Moreover, HiHa can achieve error of 1e-5 compared to other methods, surpassing their limitations, with a compression ratio of up to 27$\times$ within minutes.

Overall, the contribution of this paper are summarized in the following three aspects:
\begin{itemize}
\item The spatial characteristics are explored and analyzed in mathematical analysis, and theoretically deduce the separation criterion for harmonic;
\item A frequency-based hierarchical compression strategy is proposed, incorporating hierarchical harmonic decomposition and utilizing spherical coordinates and atmospheric pressure-level as the fundamental vectors;
\item Temporal residual module is established which contains a multiscale Laplacian Pyramid architecture and a retraining process to accelerate compression over successive periods;
\item The results of comprehensive experiments demonstrate that HiHa can achieve better compression accuracy while incurring minimal compression overhead compared with other methods.
\end{itemize}

\section{Background and Related Work}
\subsection{Atmospheric Data}

Geostationary and polar-orbiting satellites provide raw radiance data collected by ground stations to help monitor and predict weather and environmental events.
Atmospheric data contains 204 variables on over 37 atmospheric layers (known as pressure level).
Moreover, atmospheric data has a temporal dimension, composed of multiple time frames $T$.
To provide researchers with sufficient and complete global data, many institutions work to store assimilated data in transport services, such as the fifth generation ECMWF atmospheric reanalysis (ERA5) \citep{hersbach2020era5}.
Statistics from the European Centre for Medium-Range Weather Forecasts (ECMWF) show that its archive grows by about 287 terabytes (TB) on an average day \citep{schultz2021can}.

To reduce the data storage and transmission overhead, the Copernicus Atmospheric Monitoring Service (CAMS) adopts a linear quantization to compress the data, which is the most widely used GRIB2 format \citep{murrieta2015grib2} for conventional atmospheric data compression.
Moreover, the network common data form (NetCDF) \citep{rew1990netcdf}, a data abstraction for storing and retrieving multidimensional data, is proposed also as a standardized and resource-efficient approach.
These methods are the main data formats for atmospheric applications and numerical calculations, but still have a nonnegligible storage overhead.

\subsection{Implicit Neural Compression}
Based on INR, the INC is a transformative approach for representing visual information \citep{sitzmann2020implicit, 2020NeRF}.
Originally, modelling the continuous space based on coordinates, INR was proposed as a continuous and differentiable model for different types of discrete signals \citep{aghamiry2019improving}.
Theoretically, INR is highly suitable for the acquisition and compression of atmospheric data. \citep{han2024cra5}.
Specifically, the neural function can accurately describe the intensity of natural data at a voxel coordinate.
The function is defined in a continuous space, closely related to the inherent continuity of the target natural data and remains unaffected by discretization levels.
Such inherent continuity in this context is highly compatible with the spectral characteristics of dynamic equations of atmospheric motion \citep{sitzmann2020implicit}.
However, there are always trade-offs between compression ratio, compression speed and compression accuracy, due to the sophisticated multidimensional information inside the data.
To solve the contradictions, the researchers put forward innovations in the two branches of encoder-decoder and periodic activation.

\subsubsection{Encoder-decoder Compression}
The advancements in model structure works to efficiently encode atmospheric data through network structures with powerful spatially analytic capabilities.
For instance, Fourier-NN-Compression \citep{huang2022compressing} noticed the harmonic on the Earth's surface and operated a sample-based Fourier transform before feeding the data into full connection blocks, achieving outstanding performance at low resolution.
Besides, the Entroformer \citep{qian2022entroformer} and CRA5 \citep{han2024cra5} utilize Vision Transformer (ViT) Encoder as the hyper-prior model and ViT Decoder as the context model, to mitigate the computational complexity.
Moreover, C3 \citep{kim2024c3} integrates entropy coding and quantization-aware gradient-based optimization to optimize the spatial characteristics of latent grids for enhanced encoding and decoding efficiency.
In practice, such methods are more like predictions of atmospheric variables than compression, and because of the fixed encoder parameters and subdivision, the uncertainty with large distribution drift might be rapidly increased.

\subsubsection{Periodic Activation Compression}
Based on periodic activation, Siren, which adopts overfitting learning for compression, has broken through the limits of indifferentiability and model information contained in higher-order derivatives of natural signals through sine activation function, Siren \citep{sitzmann2020implicit}.
NeRF \citep{li2022nerv} proposes a new idea of multi-view reduction. The utilization of neural radiance fields as implicit expressions has proven successful in 3D point cloud reconstruction, but it exhibits limited performance in compression, akin to Gaussian Splatting \citep{zhang2024gaussianimage}.
ACORN \citep{acorn} and MINER \citep{saragadamminer} utilize multi-scale Siren network to make full use of the hierarchical data structure and characteristics.
Moreover, SCI \citep{yang2023sci} and TINC \citep{yang2023tinc} have further demonstrated the potential of periodic activation compression as continuous, memory-efficient implicit representations through combining octree storage structure with location-based sharing parameters.

The above methods combine the periodic activation function with spatial continuity and obtains great performance improvement, but their exploration and utilization of harmonics in atmospheric variables are still inadequate. 
We will theoretically explain that if combined with harmonic decomposition, there will be a broader optimization space.


\section{Mathematical Preliminary}\label{math}

We construct a difference derivation procedure from the fitting result of INR to the atmospheric variable.
An atmospheric variable $Y$ is the superposition of climatology and changing state, which is denoted by the multi-component superposition \citep{balaji2022general}:
\begin{equation}
\begin{aligned}\label{eq:MP1}
\frac{\partial Y}{\partial t}=R(X)+U(X)+P(X)+F,
\end{aligned}
\end{equation}
where $X$ denotes the space measurement, usually coordinates, $R(X)+U(X)$ denotes the Navier-Stokes equations \citep{constantin1988navier}, $P(X)$ denotes the thermodynamic process and $F$ denotes the forcing such as solar radiation.
The $\frac{\partial X}{\partial t}$ is the real information that needs to be compressed.
We use $Y$ to represent it as the over-fitting training label.
Specifically, we simplify the Eq. \ref{eq:MP1} to a polynomial wave function form:
\begin{equation}
\begin{aligned}\label{eq:MP2}
Y=\sum_0^\infty(a_i sin(\alpha_iX+\beta_i)+\delta_i)+C,
\end{aligned}
\end{equation}
where $i$ is the harmonic number, $\delta_i$ denotes the singularity, usually caused by extreme weather events, and $C$ denotes the unresolvable constant terms. 
Here we only discuss harmonics of a single time frame, so we do not use $t$ as an explicit variable.
After normalization, we eliminate the $a_i$ in Eq. \ref{eq:MP2}.

Mathematically, set $\hat{Y}$ as the INR result, we can formulate the compression task as an optimization problem
\begin{equation}
\begin{aligned}\label{eq:PM1}
min_\Theta \int_X \mathcal{L}(\hat{Y}(X,\Theta),Y)dX,
\end{aligned}
\end{equation}
where $\Theta$ denotes the paramerter set in INR. 
For decompression, one can retrieve $\widetilde{Y}_{norm}$ using storaged $\Theta$ and other meta data.
A $l$ layer Siren INR output is $sin^l(\Omega WX+B)$ \citep{sitzmann2020implicit}, where $\Omega$ denotes the hyper-parameter defined in Siren, $W$ denotes the learable parameter and $B$ denotes the Learable bias.
Therefore, the compression target is the gradient descent process, assuming $\mathcal{L}$ as the absolute difference:
\begin{equation}
\begin{aligned}\label{eq:MP3}
min \; |sin^l(\Omega WX+B)-Y|.
\end{aligned}
\end{equation}

The multi-order sine function still retains the same periodicity and parity as the first-order sine function, so if using near-linear combination as $F$, the target will turn to the higher-order difference of sine \citep{bindhu2020some}.
Let $k = \alpha_iX+\beta_i$, when meeting the condition:
\begin{equation}
\begin{aligned}\label{eq:MP4}
\Omega W \rightarrow \alpha_i, B \rightarrow \beta_i.
\end{aligned}
\end{equation}
According to the inference \citep{bindhu2020some}, Eq.\ref{eq:MP3} can be turned into
\begin{equation}
\begin{aligned}\label{eq:MP5}
\Delta_l = \lim_{n\rightarrow \infty} l|1-\frac{k^2}{2!}+...+(-1)^{(n-1)}\frac{k^n}{n!}-\sum\delta_i|+C.
\end{aligned}
\end{equation}
Using maclaurin expansion \citep{sidi2004euler}, Eq. \ref{eq:MP5} is reconstructed as
\begin{equation}
\begin{aligned}\label{eq:MP6}
\Delta_l = l|-k+e^{-k}-\sum\delta_i|+C.
\end{aligned}
\end{equation}

As the singularity, $\delta_i$ is defined as the high-frequency harmonic, and is usually regarded as noise signal.

Typically, these anomalies and singularities manifest as extreme weather events, and their specific distributions are presently difficult to generalize. 
This part will disrupt the target distribution and can be filtered out through denoising techniques that remove high-frequency signals.
Therefore, after removing $\delta_i$, we get $l|-k+e^{-k}|+C$, which has the mathematical expectation $\mathbb{E}(\Delta_l)=e^{\frac{1}{2}}+C$, obtaining $0$ when $C=-e^{\frac{1}{2}}$, demonstrating the fitability.

Siren emphasizes a special initialization $W \sim \mathcal{U}(-c,c)$, where $c$ is $\sqrt{\frac{6}{n_{c}}}$, with $n_{c}$ denoting the fan-in channel number.
Thus the harmonic with frequency $\alpha_i \in (0,\Omega \sqrt{\frac{6}{n_{c}}})$ has the highest probability of being the fastest to convergence, which we define as low-frequency harmonic.
We then set $\Omega \sqrt{\frac{6}{n_{c}}}$ as the basic frequency.
Furthermore, according to the research of El Nino and  Madden-Julian Oscillation \citep{kim2018prediction}, the first triple base frequency harmonic contains signal of special weather event, so we set $(\Omega \sqrt{\frac{6}{n_{c}}},3\Omega \sqrt{\frac{6}{n_{c}}})$ as the mid-frequency harmonic, while $(3\Omega \sqrt{\frac{6}{n_{c}}},\infty)$ as the high-frequency harmonic.

Besides, the atmospheric variables conform to the thermal circulation in Eq. \ref{eq:MP1}, so the five basic atmospheric variables ($\mathcal{Z}$ for geopotential, $\mathcal{Q}$ for humidity, $\mathcal{T}$ for temperature, $\mathcal{U}$ for u-wind and $\mathcal{V}$ for v-wind) with other extended forms also conform the above inferences..
Therefore, atmospheric data compression tasks can adopt a unified scheme for multi-dimensional and multi-modal data.

\section{Proposed Methods}
\begin{figure*}[t]
	\centering
	\includegraphics[width=1.0\textwidth]{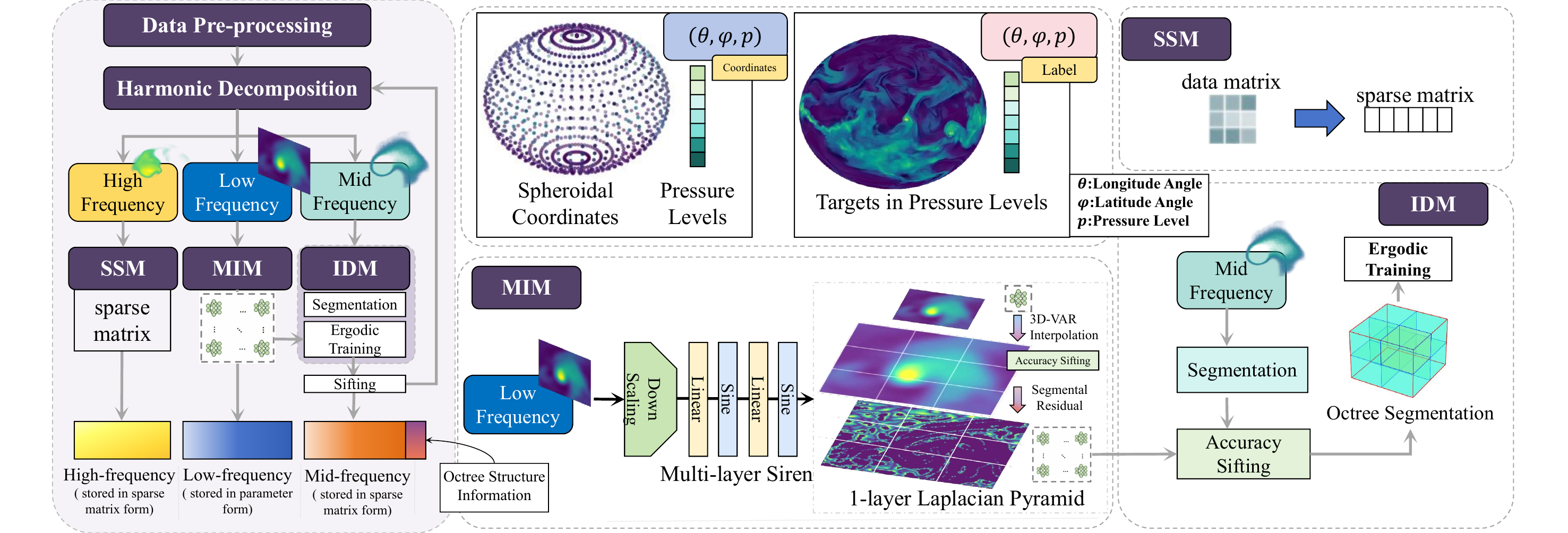}
	\caption{Flowchart of Frequency-based Iterative Compression, the spatial compression module in HiHa (Single Time Frame).}
	\label{flow}
 \vspace{-8pt}
\end{figure*}

HiHa is consisted of a spatial compression module as main framework for a single time frame, and a temporal compression module to accelerate the compression speed in the temporal dimension.
This section will focus on the characteristics of spatial and temporal compression, and explain how they interact to achieve both efficient and accurate performance.

\subsection{Pre-processing}
The climate state must be subtracted before processing the atmospheric data, according to the National Centers for Environmental Prediction (NCEP) standards \citep{hersbach2020era5}. 
We use $30$ years (1990-2020) average value as climatology, defined as $\overline{c}$, and the input atmospheric signal is initialized to $\widetilde{Y}=Y-\overline{c}$ in pre-processing shown in step 1 of Fig. \ref{flow}. 
After removing the climatology, $\widetilde{Y}$ is normalized to $\widetilde{Y}_{norm}$ to ensure that input interval is $[-1,1]$.

We use spherical coordinate $(\theta, \phi)$ and atmospheric pressure level $p$ to express the data, to ensure that the basic dimension of the data is three-dimensional.
The spherical coordinates are converted to 3D Cartesian coordinates on the unit sphere that $x=cos(\theta)cos(\phi)$, $y=cos(\theta)sin(\phi)$ and $z=sin(\theta)$, with the dimension of pressure level $p$, as depicted in Fig. \ref{flow}.

\subsection{Harmonic Decomposition}
As shown in step 2 in Fig. \ref{flow}, we use the 3D fast fourier transform to perform harmonic decomposition for $\widetilde{X}_{norm}$.
We set $(0,\Omega \sqrt{\frac{6}{n_{c}}})$ as the threshold frequency for low frequency harmonic, $(\Omega \sqrt{\frac{6}{n_{c}}},3\Omega \sqrt{\frac{6}{n_{c}}})$ as the threshold frequency for mid frequency harmonic, and $(3\Omega \sqrt{\frac{6}{n_{c}}},\infty)$ as the threshold frequency for high frequency harmonic, respectively, according to the deduction in the previous section.
Due to the correlation with the decomposition threshold, the hyper-parameters $\Omega$ and $n_c$ require delicately customized.
Therefore, we utilized ten years of data to explore best harmonic decomposition hyper-parameters, then calculated the average that $\Omega = 14, n_c=256$ for initialization phase, and $\Omega=22, n_c=128$ for re-decomposition phase.

\subsection{Frequency-based Hierarchical Compression Strategy}
We propose \textbf{F}requency-based \textbf{I}terative \textbf{C}ompression Strategy (\textbf{FIC}) as the spatial compression module of one time frame.
FIC consists of three sub-modules corresponding to the high, mid and low frequency harmonic signals, and they work in a hierarchical structure, ensuring independence within modules and fostering collaboration between modules.

\subsubsection{Sparse Storage Module}

We design \textbf{S}parse \textbf{S}torage \textbf{M}odule (\textbf{SSM}) for high-frequency signals.
The high-frequency signals that with a large impact to the data are usually with both high amplitude and high frequency, which occupy only a very tiny proportion of all signals.
However, this kind of signal will cause distribution shift in the global data distribution and reduce the performance and accuracy of INC.
Therefore, it is necessary to sacrifice some INR continuity in exchange for improved performance and accuracy.
As depicted in step 3 of Fig. \ref{flow}, we use sparse matrix storage to convert high-frequency signals whose amplitude meets a certain threshold into Compressed Sparse Row (CSR) data with lower storage overhead.
This transformation effectively reduces the storage overhead and minimizes the storage overhead for high-frequency signals.

\subsubsection{Multi-scale INR Module}

We design \textbf{M}ulti-scale \textbf{I}NR \textbf{M}odule (MIM) for low-frequency signals.
The low-frequency harmonics fit fastest in theory, and is the most widespread existence in the global space of atmospheric data, and exhibit little amplitude variations in global space.
However, this will bring huge memory occupation, computing overhead and straggler parameter.

Therefore, we adopt a multi-scale INC manner, utilizing an 1-layer Laplacian Pyramid structure to greatly reduce the computational overhead.
Specifically, for low-frequency harmonics, we firstly downscale the target $\widetilde{Y}_{norm}$ to a thumbnail $\widetilde{Y}_{norm}^{ds}$, which is downscaled to $\frac{1}{N_{scale}}\times$, the 4.1 step of Fig. \ref{flow}.

Then we implement the Siren based INR with $\widetilde{Y}_{norm}^{ds}$, and achieve $\widehat{Y}^{ds}_{norm}(low)$. 
Next, a 3-dimensional variational assimilation (3D-VAR) \citep{1998The} based interpolation is implemented on $\widehat{Y}^{ds}_{norm}(low)$ upscaling to $N_{scale}\times$, due to its widely application in interpolation of spatial atmospheric data, and achieves $\widehat{Y}_{norm}(low)$.

Moreover, we segment the $\widehat{Y}_{norm}(low)$ to $N_{scale}\times$ blocks, and sift out the blocks, whose residual $Re(\widehat{Y}_{norm}(low))$ between the block and the ground truth is below the target accuracy, to operates another INR to generate $Re^*(\widehat{Y}_{norm}(low))$, the compressed residual.
We set $N_{scale}$ to 27 ($3^3$) due to that the earth is composed of a tropical zone around the equator and two cold zones around the poles.
The results produced by this process are saved in the neural network parameters form.

\subsubsection{Iterative Decomposition Module}

We design \textbf{I}terative \textbf{D}ecomposition \textbf{M}odule (\textbf{IDM}) for mid-frequency signals.
Compared with other signals, the training of mid-frequency signals is the most complex task due to the following challenges:
(i) The mid-frequency signals still contain a large number of sophisticated superimposed signals inside, which increases the difficulty of INR trainings;
(ii) The mid-frequency signals exhibit localized spatial characteristics, posing challenges in globally fitting INR parameters through learning.

Firstly, in order to extract local features, the coordinate space $(\theta,\phi,p)$ is partitioned into eight blocks as octree to establish the local space.
However, to ergodicly train all these blocks will lead to lengthy INR trainings due to too coarse spatial granularity, and wasting storage space due to duplicate parameters.
Therefore, the existing combination of low frequency and high frequency signals is used to selectively sift blocks based on their accuracy. 
Insufficiently precise blocks will undergo INR trainings in order to minimize the resource overhead of compression.
When a satisfactory block cannot be trained within a specific number of steps, the harmonic decomposition of this block will be performed, and start from the step 2 in this block. 
In this way, an iterative harmonic decomposition INR is formed, and the depth structure citepp by compression and redecomposition is stored by octree structure.

\subsection{Temporal Residual Compression Module}
\begin{figure}[t]
	\centering
	\includegraphics[width=0.45\textwidth]{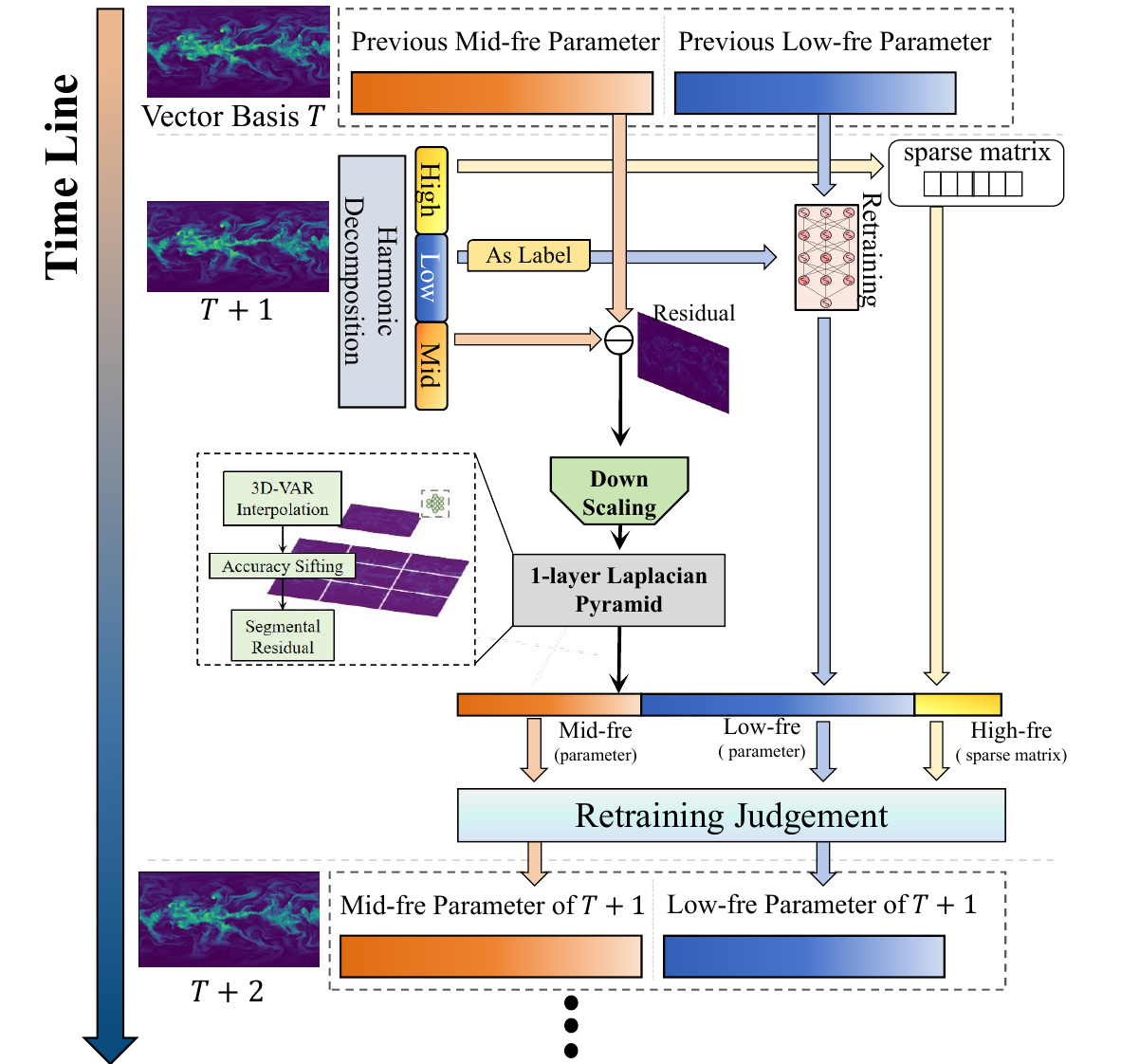}
	\caption{Flowchart of Temporal Residual Compression.}
	\label{time}
 \vspace{-10pt}
\end{figure}
In order to mine more beneficial features, we further explore the potential of INC on the temporal dimension of atmospheric data.
We design a temporal residual module, \textbf{T}emporal \textbf{R}esidual \textbf{C}ompression (\textbf{TRC}), reusing the low-frequency residuals and mid-frequency parameter of previous time frame $T$, the vector basis $\widehat{Y}(T)$.
At the next time frame $T+1$, we also operate harmonic decomposition firstly to generate high, mid and low frequency harmonics.
Due to the high complexity and irregularity, the high frequency harmonic is stored in sparse matrix form, as shown in step 1 of Fig. \ref{time}.
The changes in low-frequency signals between adjacent time frames are tiny, thus incorporating the low-frequency signal parameters from the previous time frame into the INR model for further training can achieve high accuracy with little cost.

The mid frequency signal requires fitting the residuals between adjacent time frames.
Specifically, we calculate the residual by $Re = \widetilde{Y}^{mid}_{norm}(T+1) - \widehat{Y}^{mid}_{norm}(T)$, and downscale to $Re^{ds}$.
Next, we utilize a 1-layer Laplacian pyramid, same as the Multi-scale INR Module, in that the residual of the mid frequency signal of two adjacent time frames is close to the low-frequency signal.
We firstly down scale and generate compressed $Re^{*ds}$ with INR, then deploy 3D-VAR to generate $\widehat{Re}$, accuracy sifting, and finally generate segment $Re^*(\widehat{Re})$ using INR.

After the processing of the high-, mid- and low-frequency harmonics, HiHa generate a new compression of preliminary data $T+1$.
To prevent the chaotic effect caused by error accumulation, the retraining judgment will determine whether the current compression meets the accuracy threshold or if retraining is required using spatial compression (frequency-based hierarchical compression strategy) as mentioned above. 
The entire time series data is processed, and through compressing only the residual part, compression time are significantly improved with guaranteed.

\section{Experimental Results}
\begin{table*}[htbp]
\centering
\caption{Performance Comparison of best PSNR and RMSE}
\label{comtable}
\setlength{\tabcolsep}{0.4mm}{
\begin{tabular}{c|c|ccccccccccc}
\hline
Variables                         & Metrics & HiHa(ours)                            & Entroformer                  & CRA5                          & C3                           & Fourier                       & Siren                        & NeRF                         & SCI                          & TINC                         & ACORN                        & MINER   \\ \hline
\multirow{2}{*}{$\mathcal{Z}500$} & PSNR    & \multicolumn{1}{c|}{\textbf{51.7}}    & \multicolumn{1}{c|}{13.1}    & \multicolumn{1}{c|}{31.1}     & \multicolumn{1}{c|}{21.6}    & \multicolumn{1}{c|}{27.1}     & \multicolumn{1}{c|}{15.4}    & \multicolumn{1}{c|}{16.3}    & \multicolumn{1}{c|}{22.6}    & \multicolumn{1}{c|}{22.4}    & \multicolumn{1}{c|}{19.5}    & 19.5    \\
                                  & RMSE    & \multicolumn{1}{c|}{\textbf{11.44}}   & \multicolumn{1}{c|}{284.32}  & \multicolumn{1}{c|}{33.13}    & \multicolumn{1}{c|}{56.45}   & \multicolumn{1}{c|}{296.31}   & \multicolumn{1}{c|}{229.31}  & \multicolumn{1}{c|}{231.24}  & \multicolumn{1}{c|}{59.89}   & \multicolumn{1}{c|}{72.7488} & \multicolumn{1}{c|}{96.49}   & 96.53   \\ \hline
\multirow{2}{*}{$\mathcal{Q}850$} & PSNR    & \multicolumn{1}{c|}{\textbf{54.1}}    & \multicolumn{1}{c|}{12.7}    & \multicolumn{1}{c|}{35.2}     & \multicolumn{1}{c|}{22.4}    & \multicolumn{1}{c|}{25.3}     & \multicolumn{1}{c|}{16.1}    & \multicolumn{1}{c|}{17.8}    & \multicolumn{1}{c|}{25.9}    & \multicolumn{1}{c|}{23.5}    & \multicolumn{1}{c|}{19.2}    & 19.3    \\
                                  & RMSE    & \multicolumn{1}{c|}{\textbf{9.63e-6}} & \multicolumn{1}{c|}{6.03e-4} & \multicolumn{1}{c|}{1.85e-05} & \multicolumn{1}{c|}{1.93e-4} & \multicolumn{1}{c|}{1.40e-04} & \multicolumn{1}{c|}{2.74e-4} & \multicolumn{1}{c|}{2.83e-4} & \multicolumn{1}{c|}{1.96e-5} & \multicolumn{1}{c|}{2.63e-5} & \multicolumn{1}{c|}{4.34e-5} & 3.57e-5 \\ \hline
\multirow{2}{*}{$\mathcal{T}850$} & PSNR    & \multicolumn{1}{c|}{\textbf{52.8}}    & \multicolumn{1}{c|}{13.3}    & \multicolumn{1}{c|}{37.1}     & \multicolumn{1}{c|}{19.8}    & \multicolumn{1}{c|}{26.5}     & \multicolumn{1}{c|}{14.5}    & \multicolumn{1}{c|}{15.6}    & \multicolumn{1}{c|}{24.8}    & \multicolumn{1}{c|}{23.3}    & \multicolumn{1}{c|}{19.4}    & 19.8    \\
                                  & RMSE    & \multicolumn{1}{c|}{\textbf{0.05}}    & \multicolumn{1}{c|}{1.40}    & \multicolumn{1}{c|}{0.28}     & \multicolumn{1}{c|}{0.14}    & \multicolumn{1}{c|}{1.49}     & \multicolumn{1}{c|}{1.03}    & \multicolumn{1}{c|}{1.10}    & \multicolumn{1}{c|}{0.14}    & \multicolumn{1}{c|}{0.16}    & \multicolumn{1}{c|}{0.18}    & 0.18    \\ \hline
\multirow{2}{*}{$\mathcal{T}2m$}  & PSNR    & \multicolumn{1}{c|}{\textbf{53.3}}    & \multicolumn{1}{c|}{12.8}    & \multicolumn{1}{c|}{36.3}     & \multicolumn{1}{c|}{20.5}    & \multicolumn{1}{c|}{25.8}     & \multicolumn{1}{c|}{14.3}    & \multicolumn{1}{c|}{15.3}    & \multicolumn{1}{c|}{25.5}    & \multicolumn{1}{c|}{24.7}    & \multicolumn{1}{c|}{20.1}    & 19.2    \\
                                  & RMSE    & \multicolumn{1}{c|}{\textbf{0.06}}    & \multicolumn{1}{c|}{1.89}    & \multicolumn{1}{c|}{0.67}     & \multicolumn{1}{c|}{0.27}    & \multicolumn{1}{c|}{1.37}     & \multicolumn{1}{c|}{1.23}    & \multicolumn{1}{c|}{1.71}    & \multicolumn{1}{c|}{0.47}    & \multicolumn{1}{c|}{0.46}    & \multicolumn{1}{c|}{0.38}    & 0.39    \\ \hline
\multirow{2}{*}{$\mathcal{U}10$}  & PSNR    & \multicolumn{1}{c|}{\textbf{53.9}}    & \multicolumn{1}{c|}{13.3}    & \multicolumn{1}{c|}{32.3}     & \multicolumn{1}{c|}{21.5}    & \multicolumn{1}{c|}{26.2}     & \multicolumn{1}{c|}{15.5}    & \multicolumn{1}{c|}{14.2}    & \multicolumn{1}{c|}{26.3}    & \multicolumn{1}{c|}{25.1}    & \multicolumn{1}{c|}{21.1}    & 18.1    \\
                                  & RMSE    & \multicolumn{1}{c|}{\textbf{0.09}}    & \multicolumn{1}{c|}{1.91}    & \multicolumn{1}{c|}{0.52}     & \multicolumn{1}{c|}{0.27}    & \multicolumn{1}{c|}{1.35}     & \multicolumn{1}{c|}{1.21}    & \multicolumn{1}{c|}{1.71}    & \multicolumn{1}{c|}{0.47}    & \multicolumn{1}{c|}{0.47}    & \multicolumn{1}{c|}{0.38}    & 0.39    \\ \hline
\multirow{2}{*}{$\mathcal{V}10$}  & PSNR    & \multicolumn{1}{c|}{\textbf{53.9}}    & \multicolumn{1}{c|}{13.2}    & \multicolumn{1}{c|}{32.3}     & \multicolumn{1}{c|}{21.7}    & \multicolumn{1}{c|}{26.2}     & \multicolumn{1}{c|}{15.5}    & \multicolumn{1}{c|}{14.2}    & \multicolumn{1}{c|}{26.4}    & \multicolumn{1}{c|}{25.0}    & \multicolumn{1}{c|}{20.9}    & 18.3    \\
                                  & RMSE    & \multicolumn{1}{c|}{\textbf{0.09}}    & \multicolumn{1}{c|}{1.90}    & \multicolumn{1}{c|}{0.26}     & \multicolumn{1}{c|}{0.15}    & \multicolumn{1}{c|}{1.49}     & \multicolumn{1}{c|}{1.13}    & \multicolumn{1}{c|}{1.10}    & \multicolumn{1}{c|}{0.14}    & \multicolumn{1}{c|}{0.13}    & \multicolumn{1}{c|}{0.19}    & 0.18    \\ \hline
\end{tabular}
}
\end{table*}
\subsection{Implementation}
\subsubsection{Implementation Details}
We implement HiHa with \textit{PyTorch} \citep{imambi2021pytorch} for INR trainings.
The hardware used in the experiment is Intel(R) Xeon(R) Gold 6348 @ 2.60GHz, 128GB DDR4 DRAM and NVIDIA GeForce RTX 3090.

We utilized the ERA5 \citep{hersbach2020era5} global datasets with $n_{var}$ variables, 0.25 resolution, 37 pressure levels and one year of hourly data, shaped as $n_{var}\times37\times721\times1440\times8760$, for the experiment.

Through experiments and mathematical approximation, we set 3 layers for multi-layer Siren in step 4.2 of FIC (Fig .\ref{flow}), 2 layers for residual Siren in step 4.4 of FIC, 4 layers for ergodic training in step 5.3 of FIC, 3 layers for multi-layer Siren in step 5 of TRC (Fig .\ref{time}) and 2 layers for temporal residual Siren in step 7 of TRC.
Sinusoidal frequency is set to $\Omega_{FIC4.2}=14$, $\Omega_{FIC4.4}=\Omega_{TRC5}=15$, $\Omega_{TRC7}=16$ and $\Omega_{FIC5.3}=22$, corresponding to signals with different frequencies.
Besides, the optimizer is Adam \citep{kingma2014adam}, and we adopt dynamic learning rate, which is an important element in INR trainings.
We set initial learning rate as 1e-4, and utilize \textit{CosineAnnealingLR} in pytorch as the scheduler for dynamic learning rate.
Due to that the cosine function and the sine function contained in INR are derivatives of each other, so the correlation between the cosine and the changing rate of INR is positive.
\begin{figure}[t]
	\centering
	\includegraphics[width=0.46\textwidth]{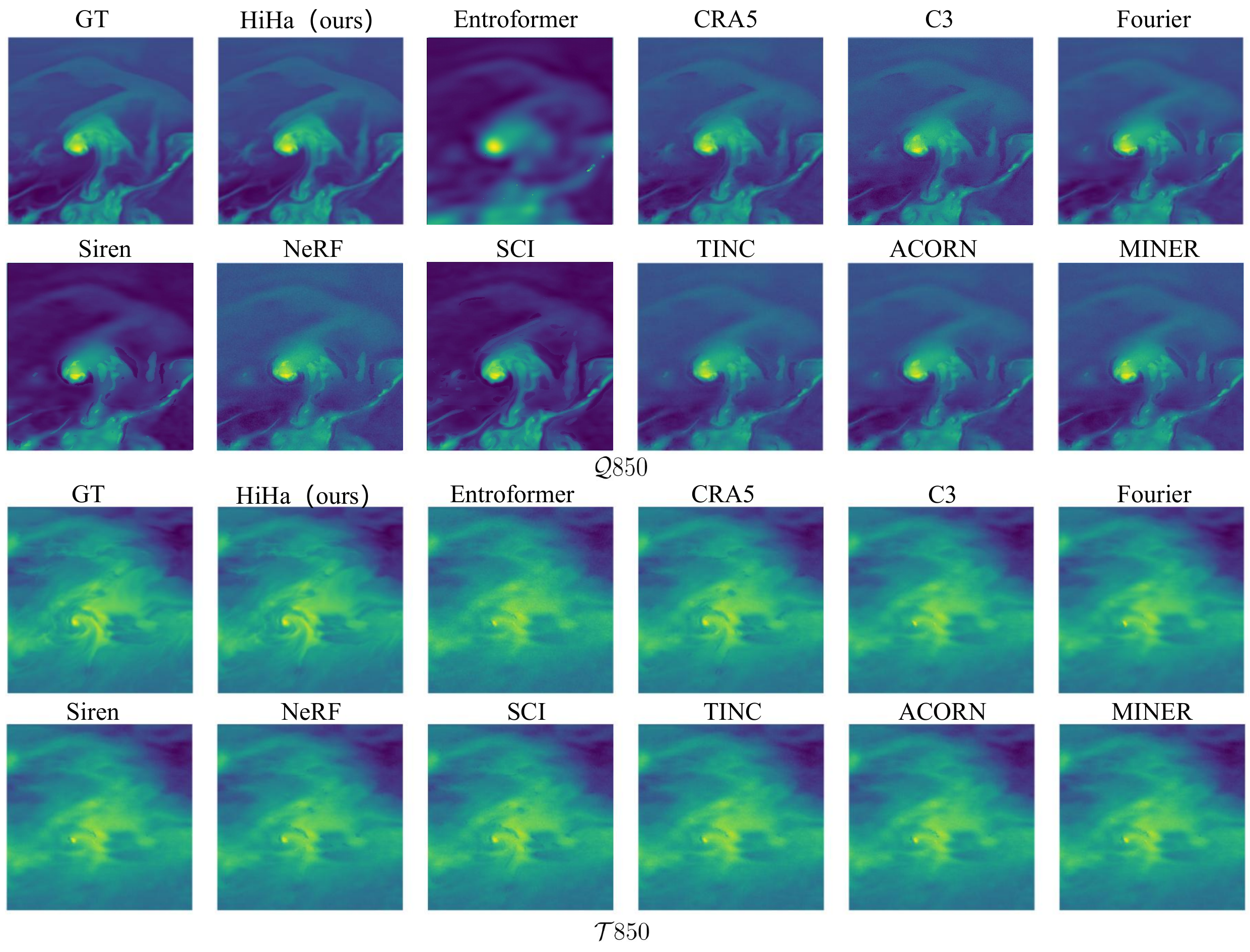}
	\caption{Visual comparison, $\mathcal{Q}850$ and $\mathcal{T}850$, where `GT' denotes the ground truth.}
	\label{viscompare}
 \vspace{-10pt}
\end{figure}

\subsubsection{Baselines and Metrics}
The proposed method is comprehensively compared with state-of-the-art methods: Grib2 \citep{murrieta2015grib2}, Netcdf \citep{rew1990netcdf}, Siren \citep{sitzmann2020implicit}, NeRF \citep{2020NeRF}, Entrofromer \citep{qian2022entroformer}, CRA5 \citep{han2024cra5}, Fourier-based method \citep{huang2022compressing}, SCI \citep{yang2023sci}, TINC \citep{yang2023tinc}, C3 \citep{kim2024c3}, ACON \citep{acorn} and MINER \citep{saragadamminer}.
For the method without temporal module, we adopt the strategy of compressing file by file.

The accuracy of INC has a limit as the compression time increasing.
We evaluate several critical metrics including best \textbf{P}eak \textbf{S}ignal to \textbf{N}oise \textbf{R}atio (PSNR/dB), best \textbf{R}oot \textbf{M}ean \textbf{S}quare \textbf{E}rror (RMSE), \textbf{B}est \textbf{C}ompression \textbf{R}ation (\textbf{BCR}) with threshold accuracy and Compression \textbf{T}ime (\textbf{$T_{com}$}) with target accuracy.
The conventional compression methods are lossless compression, so we compare PSNR and RMSE with INC methods.

We aim to present and contrast the variables that are of particular interest in atmospheric science and pose challenges for comparison, including 500hpa geopotential ($\mathcal{Z}500$), 850hpa humidity ($\mathcal{Q}850$), 850hpa temperature ($\mathcal{T}850$), surface temperature ($\mathcal{T}2m$), 10hpa U-wind ($\mathcal{U}10$) and 10hpa V-wind ($\mathcal{V}10$) for they can represent other derived variables \citep{huang2022compressing}.
\begin{table*}[h]
\centering
\caption{Compression Ratio/Time with Certain Accuracy, where \textit{-} denotes that the methods failed to reach the target accuracy or the resulting file is larger than the source file. $T_{ec}$ represents the duration of encoding training, which may extend beyond several days.}
\label{comtable2}
\setlength{\tabcolsep}{2.0mm}{
\begin{tabular}{l|llllllll}
\hline
Norm-RMSE   & \multicolumn{2}{l|}{1e-2}                            & \multicolumn{2}{l|}{1e-3}         & \multicolumn{2}{l|}{1e-4*}                           & \multicolumn{2}{l}{1e-5}            \\ \hline
Metrics     & Com Ratio            & \multicolumn{1}{l|}{Com Time} & Com Ratio            & \multicolumn{1}{l|}{Com Time} & Com Ratio           & \multicolumn{1}{l|}{Com Time} & Com Ratio           & Com Time      \\ \hline
GRIB2       & $4\times$            & 5s                 & $4\times$            & 5s                 & $4\times$           & 5s                 & $4\times$           & 5s \\
NetCDF      & $2\times$            & 5s                 & $2\times$            & 5s                 & $2\times$           & 5s                 & $2\times$           & 5s \\
Entroformer & $120\times$          & 3s+$T_{ec}$                 & \textit{-}        & \textit{-}                 & \textit{-}       & \textit{-}                 & \textit{-}       & \textit{-} \\
CRA5        & $290\times$ & \textbf{3s}+$T_{ec}$                 & $219\times$          & \textbf{3s}+$T_{ec}$                 & -          & -                 & \textit{-}       & \textit{-} \\
C3          & \textbf{479}$\times$          & 12s+$T_{ec}$                 & \textbf{360}$\times$ & 15s+$T_{ec}$                 & \textit{-}       & \textit{-}                 & \textit{-}       & \textit{-} \\
Fourier     & $308\times$          & 3s+$T_{ec}$                 & $198\times$          & 11s+$T_{ec}$                 & \textit{-}       & \textit{-}                 & \textit{-}       & \textit{-} \\
Siren       & $21\times$           & 1h                            & $5\times$            & 7h                            & \textit{-}       & -                           & \textit{-}       & -           \\
NeRF        & $21\times$           & 1h                            & $12\times$           & 5h                            & $2\times$           & 17h                           & \textit{-}       & -           \\
SCI         & $87\times$           & 41s                           & $41\times$           & 237s                          & $2\times$           & 3346s                         & \textit{-}       & \textit{-}         \\
TINC        & $79\times$           & 45s                           & $39\times$           & 356s                          & $2\times$           & 1891s                         & \textit{-}       & \textit{-}         \\
ACORN       & $63\times$           & 53s                           & $35\times$           & 3180s                         & $3\times$          & 6h                            & \textit{-}       & \textit{-}            \\
MINER       & $20\times$           & 2110s                  & $5\times$            & 763s                          & \textit{-}       & \textit{-}                         & \textit{-}       & \textit{-}           \\
HiHa-nonTRC  & $365\times$          & 18s                           & $242\times$          & 74s                  & \textbf{$67\times$} & 322s               & \textbf{$26\times$} & 937s \\ 
HiHa  & $368\times$          & 11s                           & $244\times$          & 43s                  & \textbf{68$\times$} & \textbf{183s}                 & \textbf{27$\times$} & \textbf{308s} \\ 
\hline
\end{tabular}
}
\end{table*}
\begin{table}[h]
\centering
\caption{Ablation Studies. The abbreviation `CR' represents the compression ratio, while `CT' stands for the compression time.}
\label{comtable3}
\setlength{\tabcolsep}{1.0mm}{
\begin{tabular}{c|ccc|c|c}
\hline
Items       & Variables        & PSNR  & RMSE     & CR                          & CT                     \\ \hline
\multirow{6}{*}{No SSM} & $\mathcal{Z}500$ & 44.18 & 19.13    & \multirow{6}{*}{15$\times$} & \multirow{6}{*}{3814s} \\
                        & $\mathcal{Q}850$ & 43.21 & 1.56e-05 &                             &                        \\
                        & $\mathcal{T}850$ & 44.01 & 0.10     &                             &                        \\
                        & $\mathcal{T}2m$  & 43.66 & 1.03     &                             &                        \\
                        & $\mathcal{U}10$  & 43.33 & 1.11     &                             &                        \\
                        & $\mathcal{V}10$  & 44.09 & 0.11     &                             &                        \\ \hline
\multirow{6}{*}{No MIM} & $\mathcal{Z}500$ & 40.24 & 24.88    & \multirow{6}{*}{2$\times$}  & \multirow{6}{*}{1157s}   \\
                        & $\mathcal{Q}850$ & 39.39 & 1.66e-04 &                             &                        \\
                        & $\mathcal{T}850$ & 40.33 & 0.29     &                             &                        \\
                        & $\mathcal{T}2m$  & 40.33 & 1.12     &                             &                        \\
                        & $\mathcal{U}10$  & 40.33 & 1.17     &                             &                        \\
                        & $\mathcal{V}10$  & 39.80 & 0.27     &                             &                        \\ \hline
\multirow{6}{*}{No IDM} & $\mathcal{Z}500$ & 35.15 & 31.17    & \multirow{6}{*}{50$\times$} & \multirow{6}{*}{937s}  \\
                        & $\mathcal{Q}850$ & 35.42 & 1.86e-04 &                             &                        \\
                        & $\mathcal{T}850$ & 35.12 & 0.65     &                             &                        \\
                        & $\mathcal{T}2m$  & 35.32 & 1.64     &                             &                        \\
                        & $\mathcal{U}10$  & 35.12 & 1.69     &                             &                        \\
                        & $\mathcal{V}10$  & 35.11 & 0.64     &                             &                        \\ \hline
\end{tabular}
}
\vspace{-10pt}
\end{table}

\subsection{Performance Comparison}

\subsubsection{Accuracy Performance}

The performance of these variables in compression is demonstrated in Table. \ref{comtable}. 
We also present the visual results of $\mathcal{Q}850$ and $\mathcal{T}850$ as they are the most sophisticated data for compression in Fig. \ref{viscompare}.
Unlike visual data, atmospheric data need to meet certain accuracy requirements after compression. 
Therefore, more attention based on accuracy is paid on atmospheric data compression accuracy.
The comparison focuses on both the limit accuracy and compression ratio, and the insights into the intricate relationship between compression time and these factors.

Table. \ref{comtable} demonstrates the outstanding performance of HiHa in both PSNR and RMSE compared with other INC methods.
Specifically, HiHa achieves more than 50 PSNR in all critical variables, which largely surpasses other baselines.
As we employ only one layer or a small number of INR overlays (mainly in the Laplacian pyramid), HiHa demonstrates minimal error accumulation, which is also evident in Fig \ref{viscompare}, particularly when compared to MINER.
Furthermore, the encoder-decoder structure-based methods exhibit a deficiency in precision.
Considering their high training cost, these compression methods may require more detailed matching of compressed atmospheric data time periods and variables. 
The HiHa, in contrast, exhibits enhanced flexibility and broader applicability.

The data compression quality can be visually compared with Fig \ref{viscompare}. 
A common problem in the existing methods is fuzziness, which is also a common phenomenon in the process of neural network fitting.
The harmonic decomposition-combination process effectively mitigates the issue of over-smoothing encountered in existing methods, particularly noticeable in high-frequency signals, exhibiting significant variations, such as the cyclone depicted in the figure.

\subsubsection{Compression Performance}

According to the references \citep{han2024cra5}, data with a standardized RMSE of at least 1e-4 are required to be accepted by both data-driven and numerical models.
Table. \ref{comtable2} depicts the comparison of methods including conventional methods, where the accuracy metric is the normalized RMSE, where HiHA-nonTRC denotes the HiHa without TRC module.
The time of the conventional compression method is constant, but its compression ratio is also extremely limited, which also highlights the necessity of using INC.

HiHa outperforms other methods under the requirements of high precision, in terms of compression speed and compression ratio.
Although the encoder-decoder based methods can achieve high compression ratio in a very short time after the encoder is trained, their accuracy upper bound results in the accuracy requirement being met only by CRA5.
The complex distribution of the global space necessitates a larger parameter space for periodic activation compression, rendering high-precision compression unattainable.
The variable $T_{ec}$ represents the duration of encoding training, which may extend beyond several days.
As a result, the high training costs and limited compression make these methods less efficient.

We use $*$ to mark the required accuracy, 1e-4 normalized RMSE, for both data-driven weather forecasting and numerical prediction model.
The NeRF, SCI, TINC, ACORN, HiHA-nonTRC and HiHa reach the target accuracy, which demonstrates the effectiveness in compression accuracy of INC methods.
Among these INC methods, HiHa and HiHa-nonTRC have improvement in compression accuracy and compression speed.
The hierarchical harmonic decomposition assisted with theoretical derivation significantly reduces the parameter search space, thereby reducing both the number of parameters and computational complexity.
HiHa can achieve 1e-5 normalized RMSE in about 308s, making it an efficient and suitable method to compress atmospheric data, while none of other methods can realize such target.
The experiments depict that HiHa outperforms both mainstream compressors and other INR-based methods in both compression fidelity and capabilities.

\subsection{Ablation Studies}
The comparison between HiHa and HiHa-nonTRC depicts effectiveness of temporal module that reuses the compressed parameter with high accuracy.

Fig. \ref{comtable3} shows the ablation studies of SSM, MIM and IDM for high, low and mid-frequency module without TRC, respectively.
The target accuracy is 1e-4 for normalized RMSE.
The high frequency part will have a large bias effect on the fitting process of INR. 
Thus expanding the parameter space can fit the data, which limits the compression ratio.
The sole fitting of the mid-frequency signals also requires a huge parameter space, which limits the compression ratio and compression time.
In the absence of mid-frequency signal fitting and iterative decomposition, the low-frequency signal fitting with the multi-scale INR is unable to meet the high accuracy requirements. Although it can significantly expedite the compression process, it necessitates a lengthier training period to attain adequate precision.
In summary, the collaboration among the submodules integrated in HiHa leads to a significant enhancement in performance.
\begin{figure}[tbp]
	\centering
	\includegraphics[width=0.5\textwidth]{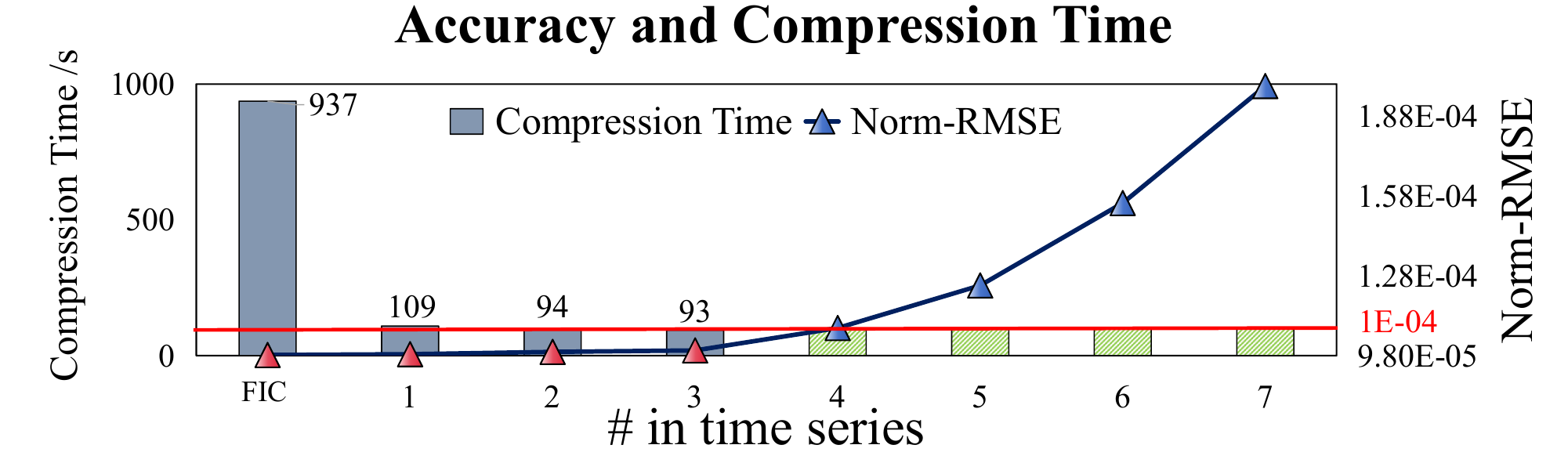}
  \vspace{-10pt}
	\caption{Variation in accuracy and compression time with increasing time series of TRC.}
	\label{ar}
 \vspace{-5pt}
\end{figure}

Fig. \ref{ar} reveals the potential error accumulation of TRC, and the corresponding benefit in speed up.
This result is obtained by averaging a large number of experiments, and the error accumulation grows nonlinearly, while the average error accumulation of 4 time steps will lead to exceeding the threshold.
The adjustment of the threshold can potentially lead to higher benefits it is crucial to carefully consider the potential adverse effects caused by error accumulation.

\subsection{DownStream: Weather Forecasting}
\begin{figure}[tbp]
	\centering
	\includegraphics[width=0.5\textwidth]{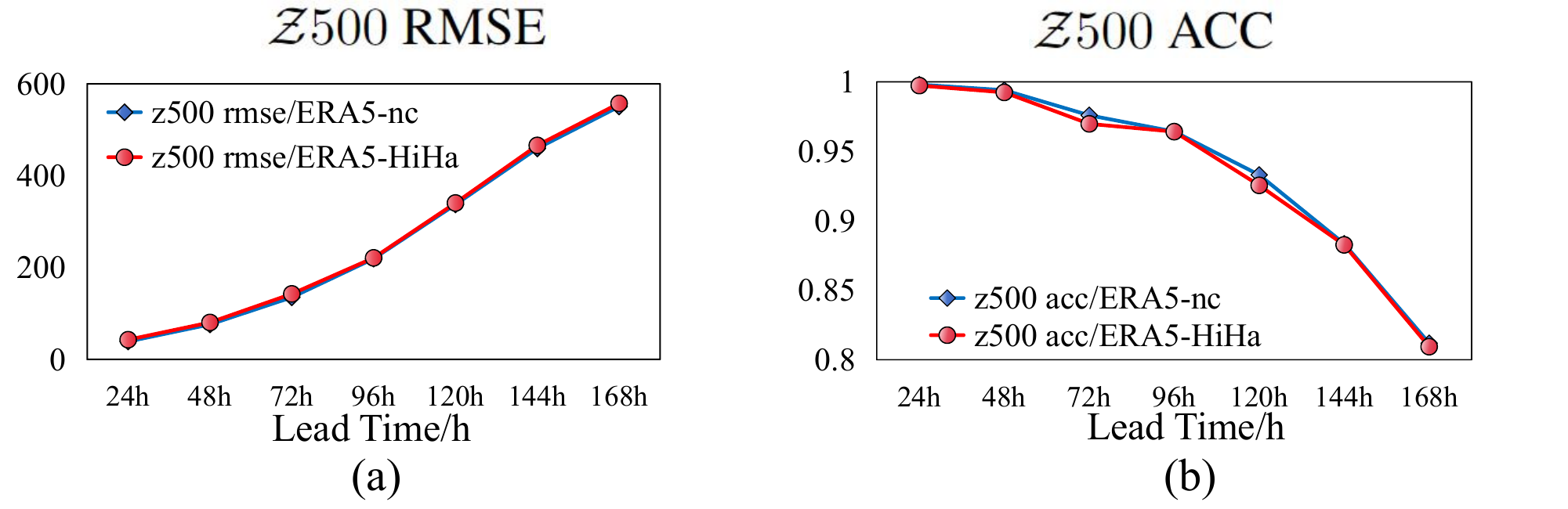}
    \vspace{-15pt}
	\caption{Accuracy Verification of $\mathcal{Z}500$ in practical applications, Pangu data-driven model \citep{bi2023accurate}.`ERA5-nc' means using data in form of netcdf (oringinal form).}
	\label{as}
 \vspace{-5pt}
\end{figure}

We need to maintain the utility of the compressed data, while pursuing a higher compression ratio.
In order to verify the effectiveness of the compressed data, we conducted a forecasting test using a data-driven weather forecasting model, as shown in Fig .\ref{as}.
We choose the representative $\mathcal{Z}500$ related results as the evaluation object.
Fig .\ref{as} (a) depicts that it is around $1\%$ deviation between the source data and compressed data from HiHa both in RMSE.
Fig .\ref{as} (b) depicts that although there is a slight difference in 72h and 120h forecasting, it has little effect on the overall results in ACC.
The deviation is deemed acceptable given the inherent corrective and fault tolerance capabilities of data-driven model.

Reducing the data storage and transmission overhead, the compressed data can still maintain the model's end-to-end prediction performance.
The experiment result suggests that HiHa compression is suitable for
scientific research.

\section{Conclusions}
This work proposed a hierarchical harmonic decomposition INC method, HiHa, for large scale atmospheric data compression. 
The method contains harmonic decomposition, a frequency-based hierarchical compression strategy, hierarchical harmonic decomposition and a temporal residual compression module.

Experiments depict that HiHa achieves outstanding performance in atmospheric data compression compared with both mainstream compressors and other INC methods in both compression fidelity and capabilities, and also demonstrate that using compressed data in existing data-driven models can achieve the same accuracy as raw data.


\bibliography{ref}
\bibliographystyle{aaai25}

\end{document}